\documentclass[journal]{IEEEtran}
\ifCLASSINFOpdf
\else
\fi

\usepackage{graphicx}
\usepackage{subfigure}
\usepackage{multirow}
\usepackage{multicol}
\usepackage{amsmath}
\usepackage{array}
\usepackage{tabulary}
\newcolumntype{K}[1]{>{\centering\arraybackslash}p{#1}}
\usepackage{threeparttable}
\usepackage{adjustbox}
\usepackage{pifont}
\newcommand{\cmark}{\ding{51}}%
\newcommand{\xmark}{\ding{55}}%
\usepackage{diagbox}
\usepackage[dvipsnames]{xcolor}

\usepackage{url}
\usepackage{mathrsfs,amsfonts,amsthm,amssymb,algorithm}

\usepackage{algorithm}
\usepackage{algorithmicx}
\usepackage{algpseudocode}

\begin{document}
%
\title{Dual Learning for Semi-Supervised Natural Language Understanding}
%
%
%

\author{Su~Zhu,~\IEEEmembership{Student~Member,~IEEE,} Ruisheng~Cao, and~Kai~Yu,~\IEEEmembership{Senior~Member,~IEEE}
\thanks{
Su Zhu, Ruisheng Cao and Kai Yu are supported by the National Key Research and Development Program of China (Grant No.2017YFB1002102). Experiments have been carried out on the PI supercomputer at Shanghai Jiao Tong University. (\emph{Su Zhu and Ruisheng Cao contribute equally to this article.}) (\emph{Corresponding authors}: Kai Yu.)

The authors are with the SpeechLab, Department of Computer Science and Engineering, and MoE Key Lab of Artificial Intelligence, AI Institute, Shanghai Jiao Tong University, Shanghai 200240, China. (e-mail: paul2204@sjtu.edu.cn; 211314@sjtu.edu.cn; kai.yu@sjtu.edu.cn)
}
}

\maketitle


\begin{abstract}
Natural language understanding (NLU) converts sentences into structured semantic forms. The paucity of annotated training samples is still a fundamental challenge of NLU. To solve this data sparsity problem, previous work based on semi-supervised learning mainly focuses on exploiting unlabeled sentences. In this work, we introduce a dual task of NLU, semantic-to-sentence generation (SSG), and propose a new framework for semi-supervised NLU with the corresponding dual model. The framework is composed of dual pseudo-labeling and dual learning method, which enables an NLU model to make full use of data (labeled and unlabeled) through a closed-loop of the primal and dual tasks. By incorporating the dual task, the framework can exploit pure semantic forms as well as unlabeled sentences, and further improve the NLU and SSG models iteratively in the closed-loop. The proposed approaches are evaluated on two public datasets (ATIS and SNIPS). Experiments in the semi-supervised setting show that our methods can outperform various baselines significantly, and extensive ablation studies are conducted to verify the effectiveness of our framework. Finally, our method can also achieve the state-of-the-art performance on the two datasets in the supervised setting.  Our code is available at \url{https://github.com/rhythmcao/slu-dual-learning.git}.
\end{abstract}

%
\IEEEpeerreviewmaketitle

\section{Introduction}
\label{sec:intro}
%
%
%
%

\IEEEPARstart{R}{ecently}, the development of mobile internet and smart devices has led to the tremendous growth of conversational dialogue systems, such as Amazon Alexa, Google Assistant, Apple Siri, and Microsoft Cortana. Natural language understanding (NLU) is a key component of these systems, parsing user's utterances into the corresponding semantic forms~\cite{wang2005spoken} for certain narrow domain (e.g., \emph{booking hotel}, \emph{searching flight}). Typically, the primary task of the NLU module in goal-oriented dialogue systems usually contains two sub-tasks: intent detection and slot filling~\cite{wang2005spoken,xu2013convolutional,liu2016attention,ZhangA,goo-etal-2018-slot,chen2019bert,qin-etal-2019-stack}. The intent detection is typically treated as a sentence classification problem~\cite{de2008spoken,tur2011intent,sarikaya2014application}, while the slot filling is typically treated as a sequence labeling problem in which contiguous sequences of words are tagged with semantic labels (slots)~\cite{wang2011semantic,he2003data,raymond2007generative,mesnil2015using}.

Deep learning has achieved great success for the intent detection and slot filling in NLU~\cite{xu2013convolutional,liu2016attention,ZhangA,goo-etal-2018-slot,chen2019bert,qin-etal-2019-stack,mesnil2015using,yao2014spoken,vu2016sequential,kurata2016leveraging,zhu2016encoder,li2018self}, outperforming most traditional approaches~\cite{raymond2007generative,Zettlemoyer2007Online} in the field of supervised learning. However, the deep learning method is notorious for requiring large labeled data, which limits the scalability of NLU models to new domains due to the annotation cost. Semi-supervised learning methods are adopted to solve this data sparsity problem of NLU, which utilize a large number of unannotated sentences to enhance the supervised NLU training~\cite{tur2005combining,lee2013pseudo,celikyilmaz2016new,oyl11-lan-icassp18}. These semi-supervised learning methods focus on exploiting the unlabeled sentences to enhance input encoders or create additional samples with predicted pseudo-labels.

Apart from \emph{pure} sentences (i.e., unannotated sentences), \emph{pure} semantic forms (i.e., intents and slots without sentence expressions) can also be utilized in the semi-supervised NLU. Exploiting semantic forms could be more affordable and effective than collecting in-domain sentences, since they are well-structured and could be automatically created or synthesized under domain knowledge. However, the previous methods of semi-supervised NLU cannot utilize pure semantic forms data.


\begin{figure}[t] 
\centering
\includegraphics[width=0.42\textwidth]{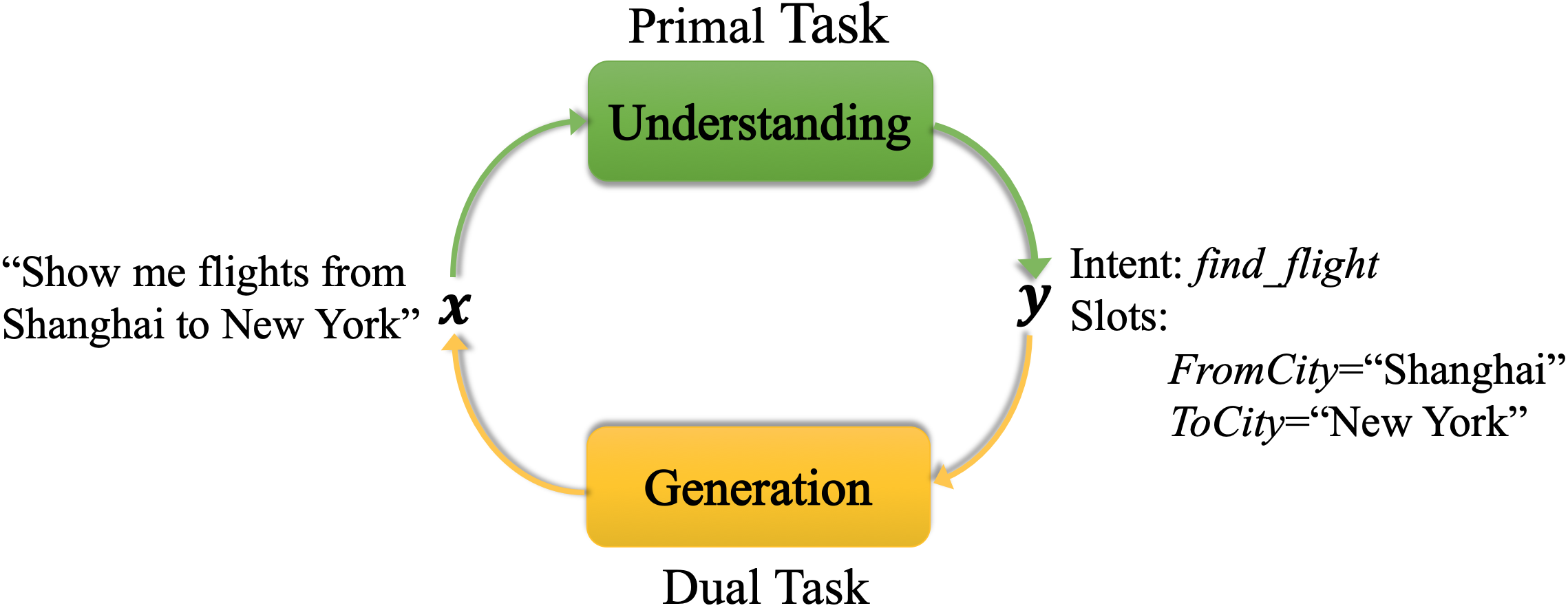}
\caption{A diagram of NLU and its dual task. The primal task is NLU, which converts an input sentence into the corresponding intent and slots. In the inverse direction, the dual task is semantic-to-sentence generation (SSG), which converts the intent and slots into a natural language sentence.}
\label{fig:dual_task}
\end{figure}

In this work, we introduce the dual task of intent detection and slot filling in NLU, as shown in Fig. \ref{fig:dual_task}. By incorporating the dual task, 
a novel framework of semi-supervised NLU is proposed, which can utilize not only pure sentences but also pure semantic forms (i.e., intents and slots). Our framework consists of two parts: a dual pseudo-labeling method and dual learning algorithm. 1) Besides using a primal model to generate pseudo labels~\cite{lee2013pseudo} for unlabeled sentences, the dual pseudo-labeling method also utilizes a dual model to generate pseudo sentences for pure semantic forms. Next, we combine these pseudo-labeled samples with the labeled dataset to retrain both the primal and dual models iteratively. 2) Furthermore, the dual learning algorithm ~\cite{he2016dual} is applied to train the primal and dual models jointly in a closed-loop of the two models. New validity rewards are proposed to validate potential sentences and semantic forms.




The main contributions of this paper are summarized:
\begin{itemize}
    \item A dual model for joint intent detection and slot filling in NLU is introduced to generate sentences based on structured semantic forms.
    \item We propose a novel framework for semi-supervised NLU by incorporating the dual model, which can better utilize unlabeled data. 
    \item We present extensive experiments on ATIS~\cite{hemphill1990atis} and SNIPS~\cite{2018arXiv180510190C} datasets, which demonstrate the benefit of our proposed framework for semi-supervised NLU. It also achieves the state-of-the-art performance in the supervised setting. 
\end{itemize}


The rest of the paper is organized as follows. The following section discusses related works. We introduce the intent detection and slot filling in NLU in Section \ref{sec:st_slu}, then describe the details of the dual task in Section \ref{sec:dual_task_of_NLU}. A semi-supervised NLU framework with the dual task is proposed in Section \ref{sec:dual_semi_supervised_NLU}. Detailed experimental results and analysis are given in Section \ref{sec:exp}. Section \ref{sec:conclusion} summarizes this work and the future direction.


\section{Related Work}

This section describes previous literature of intent detection and slot filling in NLU as well as the semi-supervised NLU.

\subsection{Intent Detection and Slot Filling in NLU}

Recently, motivated by a number of successful neural network and deep learning methods in natural language processing, many neural network architectures have been applied in the intent detection and slot filling, such as vanilla recurrent neural network (RNN) ~\cite{mesnil2015using,mesnil2013investigation,yao2013recurrent,vu2016bi}, convolutional neural network (CNN) ~\cite{vu2016sequential,xu2013convolutional,Chen2016ZeroshotLO}, long short-term memory (LSTM)~\cite{yao2014spoken,ZhangA,hakanni-tur2016multidomain,reimers2017optimal}, encoder-decoder~\cite{zhu2016encoder,liu2016attention,zhai2017neural,kurata2016leveraging}, capsule neural networks~\cite{zhang2019joint}, transformers~\cite{zhang2019using}, etc. Several pre-trained language models are also applied to improve generalization, like ELMo~\cite{Aditya2018Unsupervised} and BERT~\cite{chen2019bert,castellucci2019multi}. Most of the previous work tends to share the encoders of the intent detection and slot filling while leaves their decoders (e.g., classification layers) independent. Besides, some investigations focus on interrelated modeling of intent detection and slot filling ~\cite{li2018self,goo-etal-2018-slot,qin-etal-2019-stack,haihong2019novel}, which is orthogonal to the semi-supervised learning of NLU.

\subsection{Semi-supervised NLU}

The traditional approaches of semi-supervised NLU utilize unlabeled sentences to improve NLU performances in two ways. 1) NLU model trained with the existing labeled sentences is exploited to predict pseudo-labels for unlabeled sentences, which can be used to retrain the NLU model~\cite{tur2005combining,lee2013pseudo,celikyilmaz2016new}. 2) Except for the pseudo-labeling method, some prior works design several unsupervised tasks to make use of the unlabeled sentences, like language models~\cite{rei2017semi,peters2017semi,oyl11-lan-icassp18,Aditya2018Unsupervised}, sequence-to-sequence based sentence reconstruction~\cite{kim2017adversarial,zhu2018robust}. They share partial parameters between the unsupervised tasks and the NLU task. However, we are the first to exploit pure semantic forms (without sentence expressions) by developing a dual pseudo-labeling method.

The dual learning algorithm is first proposed for neural machine translation~\cite{he2016dual}, where translation from the target language to source language (i.e., back-translation) is the dual task. The dual learning is also applied in semantic parsing~\cite{cao2019semantic,ye2019jointly} and natural language understanding~\cite{su2019dual,su2020towards}. Su \emph{et al}.~\cite{su2019dual} propose a dual supervised learning method for natural language understanding and generation. However, their method is not compatible with a semi-supervised problem. Su \emph{et al}.~\cite{su2020towards} also train NLU and NLG (natural language generation) modules jointly and forms two closed training loops to flow the gradients, which introduces few tricks to allow gradients flow through the model chain. However, we focus on the NLU task by considering NLG as an auxiliary task, and flow the gradients with policy gradient based reinforcement learning and compact rewards. Moreover, they simplify the NLU task into a multi-label classification problem, which is not scalable. We are the first to propose a dual task for intent detection and slot filling in NLU and utilize the dual task in semi-supervised NLU.

\section{Intent Detection and Slot Filling in NLU}
\label{sec:st_slu}

This section introduces the NLU task and describes the basic multitask framework of intent detection and slot filling.

\subsection{NLU Task Formulation}

\begin{figure}[htbp]
\centering
\includegraphics[width=0.4\textwidth]{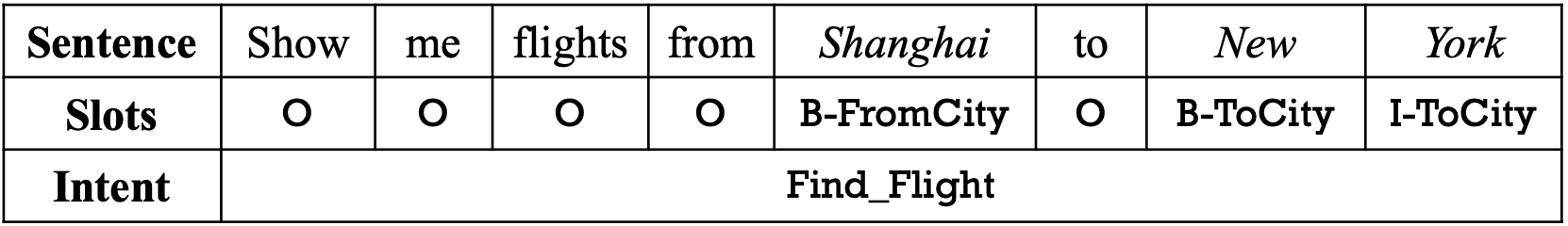}
\caption{An example of intent and slot annotation (IOB format) in ATIS dataset.}
\label{fig:atis}
\end{figure}

Intent detection and slot filling are major tasks of NLU in task-oriented dialogue systems. An intent is a purpose or a goal that underlies a user-generated utterance~\cite{xia2018zero}. Therefore, intent detection can be seen as a classification
problem to determine the intent label of an input sentence. Slot filling aims to automatically extract a set of attributes or ``slots'', with the corresponding values. It is typically treated as a sequence labeling problem. An example of data annotation is provided in Fig. \ref{fig:atis}. The user's intent is to find flights. For slot annotation, it follows the popular inside/outside/beginning (IOB) schema, where \emph{Boston} and \emph{New York} are the departure and arrival cities specified as the slot values in the user's utterance, respectively. In this work, we use the word \emph{tag} as an alias for \emph{slot} to denote semantic labels in IOB schema.

Let $x = (x_1, \cdots, x_{|x|})$ denote an input sentence (word sequence), $o^I$ denote its intent label, and $o^S = (o^S_1, \cdots, o^S_{|x|}$) denote its output sequence of slot tags, where $|x|$ is the sequence length. Each $o^S_i \in \mathcal{T}$ and $o^I \in \mathcal{I}$, where $\mathcal{T}$ and $\mathcal{I}$ are the sets of all possible slot tags and intent labels respectively in the current domain. Therefore, the intent detection and slot filling in NLU are to estimate $p(o^I, o^S|x)$, the joint posterior probability of intent $o^I$ and slot sequence $o^S$ given input $x$. Usually, the two sub-tasks are modelled independently, i.e.,
\begin{equation}
p(\widetilde{y}|x)=p(o^I, o^S|x)=p(o^I|x)p(o^S|x)
\end{equation}
where $\widetilde{y}=(o^I, o^S)$.

\subsection{Preliminaries for Neural Network}
Before providing details of the NLU model, we first introduce two basic NN modules for conciseness.

\textbf{BLSTM:} As mentioned before, many neural network architectures have been applied in intent detection and slot filling tasks. In this paper, bi-directional LSTM based RNN (BLSTM) is adopted for sequence encoding. Given a sequence of feature vectors ($\mathbf{e}_1,\cdots,\mathbf{e}_{L}$), hidden vectors are recursively computed at the $i$-th time step ($i\in\{1,\cdots,L\}$) via
\begin{equation}
    \overrightarrow{\mathbf{h}}_i=\text{f}_\text{{LSTM}}(\mathbf{e}_i, \overrightarrow{\mathbf{h}}_{i-1}); \overleftarrow{\mathbf{h}}_i=\text{f}_\text{{LSTM}}(\mathbf{e}_i, \overleftarrow{\mathbf{h}}_{i+1})
\end{equation}
and $\mathbf{h}_i=\overrightarrow{\mathbf{h}}_i\oplus\overleftarrow{\mathbf{h}}_i$, where $\oplus$ denotes the vector concatenation and $\text{f}_\text{{LSTM}}$ is the LSTM function. For convenience, we rewrite the entire operation as a mapping $\text{BLSTM}_{\Theta}$:
\begin{equation}
    (\mathbf{h}_1,\cdots,\mathbf{h}_{L}) \leftarrow \text{BLSTM}_{\Theta}(\mathbf{e}_1,\cdots,\mathbf{e}_{L}) .
\end{equation}

\textbf{Attention Mechanism:} Attention mechanism~\cite{bahdanau2014neural,luong2015effective} is usually used to obtain a sequence-level feature vector or context representations in encoder-decoder architectures. Given a sequence of feature vectors ($\mathbf{e}_1,\cdots,\mathbf{e}_{L}$) and a query vector $\mathbf{q}$, the attention weight for $\mathbf{q}$ with each $\mathbf{e}_i$ ($i\in\{1,\cdots,L\}$) is $a_i = \text{exp}(u_i) / \sum_{j=1}^{L}\text{exp}(u_j)$, and
\begin{equation}
    u_i = \mathbf{v}^{\top}_a\text{tanh}(\mathbf{W}_{a}(\mathbf{q} \oplus \mathbf{e}_i))
\end{equation}
where $\mathbf{v}_a$ and $\mathbf{W}_a$ are learnable parameters. Finally, a context vector is computed as $\mathbf{z}=\sum_{i=1}^{L} a_i \mathbf{e}_i$. For brevity, we rewrite the entire operation as a mapping returning the context vector and attention weights:
\begin{equation}
    \mathbf{z}, (a_1,\cdots,a_L) \leftarrow \text{ATTN}_{\Theta}(\mathbf{q}, (\mathbf{e}_1,\cdots,\mathbf{e}_L)) .
\end{equation}


\subsection{NLU Model Architecture}

The basic multitask framework of intent detection and slot filling is comprised of three modules: sentence encoding, intent classification, and slot tagging.

\textbf{Sentence Encoding:}
Every input word is mapped to a vector via $\mathbf{x}_i=\mathbf{W}_{x}\mathbf{o}(x_i)$, where $\mathbf{W}_{x}$ is an embedding matrix and $\mathbf{o}(x_i)$ a one-hot vector. An BLSTM encoder is applied to get hidden vectors $\mathbf{h}_{i}^{\text{sen}} \in \mathbb{R}^{2n}$ ($n$ is the hidden size, $i\in\{1,\cdots,|x|\}$):
\begin{equation}
    (\mathbf{h}_1^{\text{sen}},\cdots,\mathbf{h}_{|x|}^{\text{sen}}) \leftarrow \text{BLSTM}_{\Theta_1}(\mathbf{x}_1,\cdots,\mathbf{x}_{|x|}) .
\end{equation}

\textbf{Intent Classification:} An attention model is applied to gather a sentence embedding, and then feed it into a linear output layer for intent classification:
\begin{align}
    \mathbf{z}^{\text{sen}}, (a_1^{\text{sen}},\cdots,a_{|x|}^{\text{sen}}) &\leftarrow \text{ATTN}_{\Theta_2}(\overleftarrow{\mathbf{h}}_1^{\text{sen}}, (\mathbf{h}_1^{\text{sen}},\cdots,\mathbf{h}_{|x|}^{\text{sen}}))\\
    p(o^I|x) &= \text{softmax}_{o^I}(\mathbf{W}_1 \mathbf{z}^{\text{sen}}) 
\end{align}
where $\mathbf{W}_1 \in \mathbb{R}^{|\mathcal{I}|\times 2n}$ is trainable (bias is omitted).

\textbf{Slot Tagging:} Slot filling is considered as a sequence labeling problem which tags each input word sequentially. There are three typical methods for slot tagging concerning the time series dependence of slot tags, as shown below.

\subsubsection{BLSTM-softmax}
At each time step, a linear output layer is applied to predict slot tags independently, i.e. 
\begin{equation}
p(o^S|x) = \prod_{i=1}^{|x|} p(o^S_i|\mathbf{h}_i^{\text{sen}}) = \prod_{i=1}^{|x|} \text{softmax}_{o^S_i}(\mathbf{W}_2 \mathbf{h}_i^{\text{sen}}) 
\label{eqn:blstm-softmax}
\end{equation}
where $\mathbf{W}_2 \in \mathbb{R}^{|\mathcal{T}|\times 2n}$ is trainable, and $\text{softmax}_{o^S_i}(.)$ is precisely the $o^S_i$-th element of the distribution defined by the \emph{softmax} function.


\subsubsection{BLSTM-CRF}
Conditional Random Field (CRF) output layer considers the correlations between tags in neighborhoods and jointly decode the best chain of tags for a given input sentence~\cite{yao2014recurrent,huang2015bidirectional,ma-hovy-2016-end}. The posterior probability of slot sequence is computed via:
\begin{align}
\psi(x,o^S) &= \sum_{i=1}^{|x|} ([\mathbf{A}]_{o^S_{i-1},o^S_i} + [\mathbf{W}_2 \mathbf{h}_i^{\text{sen}}]_{o^S_i})\\ 
p(o^S|x) &= \frac{
\text{exp}(\psi(x,o^S))
}{
\sum_{o^{S'}}\text{exp}(\psi(x,o^{S'}))
}
\end{align}
where $\mathbf{A} \in \mathbb{R}^{|\mathcal{T}|\times |\mathcal{T}|}$ is a transition matrix, and its element $[\mathbf{A}]_{m,n}$ models the transition from the $m$-th to the $n$-th label for a pair of consecutive time steps.

\subsubsection{BLSTM-focus}
To consider the time series dependence of slot tags, several encoder-decoder architectures~\cite{kurata2016leveraging,zhu2016encoder,liu2016attention,zhai2017neural} are also proposed for slot filling. With the focus mechanism~\cite{zhu2016encoder}, we utilize a uni-directional LSTM based decoder to model tag dependencies. The decoder’s hidden vector at the $i$-th time step is computed by $\mathbf{h}_i^{\text{tag}}=\text{f}_\text{{LSTM}}(\mathbf{h}_i^{\text{sen}}\oplus\mathbf{o}^S_{i-1}, \mathbf{h}_{i-1}^{\text{tag}})$, where $\mathbf{o}^S_{i-1}$ is the embedding of the previously predicted slot tag, and $\mathbf{h}_0^{\text{tag}}=\overleftarrow{\mathbf{h}}_1^{\text{sen}}$. Then we compute $p(o^S|x)$ via: 
\begin{equation*}
p(o^S|x) = \prod_{i=1}^{|x|} p(o^S_i|o^S_{<i}, x) = \prod_{i=1}^{|x|} \text{softmax}_{o^S_i}(\mathbf{W}_3 \mathbf{h}_i^{\text{tag}}) 
\end{equation*}
where $\mathbf{W}_3 \in \mathbb{R}^{|\mathcal{T}|\times n}$. Compared with \emph{BLSTM-CRF}, this method can model longer-range dependence of slot tags.

\textbf{Slot-Value Summary:} Though we can get a sequence of slot tags after slot tagging, it is a semifinished representation that the value for each predicted slot is not revealed. With alignment between the predicted tag sequence and input sentence, we can extract a summary of slot-value pairs easily. For instance, the list of slot-value pairs for the sample in Fig. \ref{fig:atis} is \emph{(FromCity=Shanghai, ToCity=New York)}. Let $o^C$ denote the list of slot-value pairs, and then we can get the final semantic form $y$ of the input $x$:
\begin{equation}
    y = (o^I, o^C) = (o^I, \text{getSummary}(o^S, x)) .
\end{equation}

The loss function of the NLU model given $x$ and $y$ is 
\begin{equation*}
    \mathcal{L}_{\text{NLU}}(x, y) = -\log p(\widetilde{y}|x) = -\log p(o^I|x) -\log p(o^S|x) .
\end{equation*}


\section{Semantic-to-Sentence Generation}
\label{sec:dual_task_of_NLU}


\begin{figure*}[t] 
\centering
\includegraphics[width=0.98\textwidth]{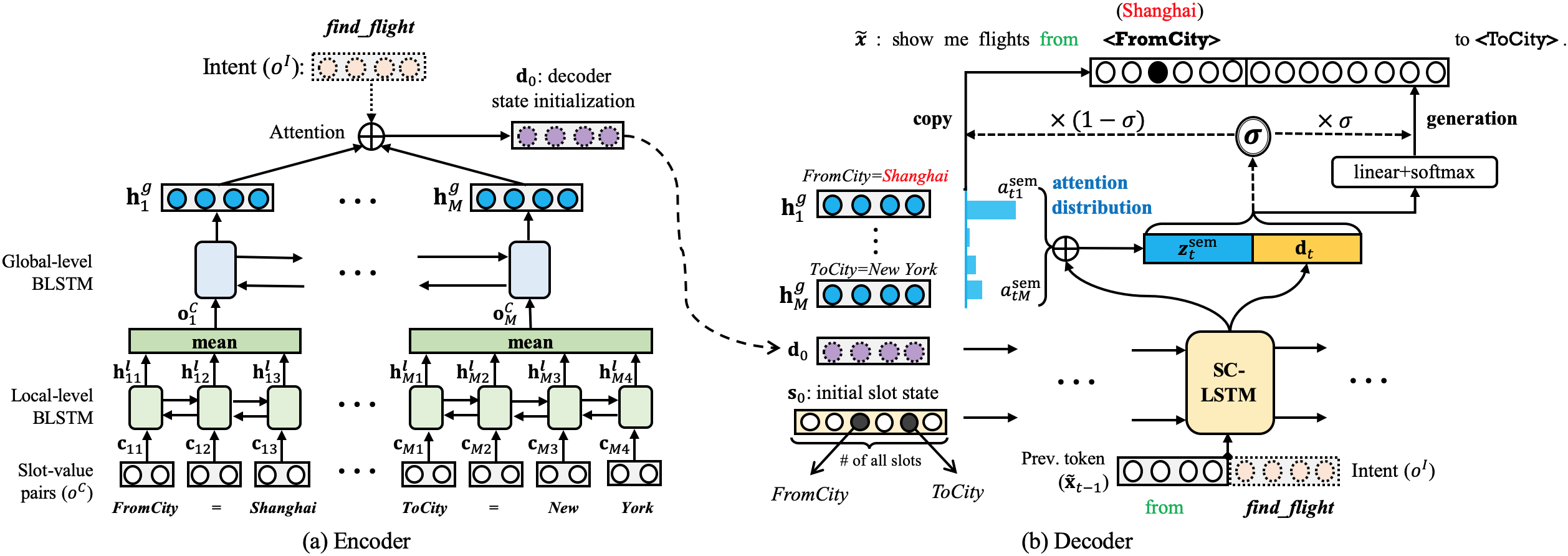}
\caption{The proposed architecture for the dual task of NLU, which is comprised of an encoder and a decoder. The encoder is a hierarchical BLSTM to obtain deep features for a list of slot-value pairs $o^C$. The decoder exploits a semantically controlled LSTM to precisely generate a delexicalized form $\widetilde{x}$, and then substitute the special slot tokens with the corresponding values in $o^C$.}
\label{fig:dual_model}
\end{figure*}

In this section, we will introduce the dual task of NLU, which is formulated as a semantic-to-sentence generation (SSG) task. It generates the corresponding sentence $x$ given an intent $o^I$ and a list of slot-value pairs $o^C=(o^C_1,\cdots,o^C_{M})$ ($M$ is the number of slot-value pairs). As a fact of the IOB annotation schema, each value in $o^C$ must appear in $x$ without overlapping. Thus, we choose to first generate a delexicalized form\footnote{For example, the delexicalized form of the sentence in Fig. \ref{fig:atis} is ``\emph{show me flights from} $\langle\texttt{FromCity}\rangle$ \emph{to} $\langle \texttt{ToCity} \rangle$''.} $\widetilde{x}$ comprised of words and slots, and then fill the slots up with given values in $o^C$ to get $x$. We wish to estimate 
\begin{equation}
    p(\widetilde{x}|y) = p(\widetilde{x}|o^I, o^C) ,
\end{equation}
the conditional probability of delexicalized form $\widetilde{x}$ given intent $o^I$ and slot-value pairs $o^C$.


Sequence-to-sequence based encoder-decoder architectures has achieved success in natural language generation such as machine translation ~\cite{bahdanau2014neural,luong2015effective}, dialogue generation~\cite{vinyals2015neural} and text summarization~\cite{see2017get}. However, it is non-trivial to apply the encoder-decoder architectures into SSG, since the input of SSG is not a sequence any more but a structured form (i.e., an intent and a list of slot-value pairs). 

The proposed architecture for SSG is illustrated in Fig. \ref{fig:dual_model}. An \emph{encoder} is exploited to encode the intent $o^I$ and the list of slot-value pairs $o^C$ into vector representations, and a \emph{decoder} learns to generate the delexicalized form $\widetilde{x}$ depending on the encoding vectors. Finally, we replace the slots in $\widetilde{x}$ with the corresponding values in $o^C$.

\textbf{Encoder:} We exploit a hierarchical BLSTM to encode the list of slot-value pairs at local and global levels. Firstly, each slot-value pair is considered as a sub-sequence, i.e., $o^C_m=(c_{m1},\cdots,c_{mT_m})$, where $T_m$ is the sequence length, and $m \in\{1,\cdots,M\}$. For example, a slot-value pair, \emph{ToCity=New York}, is tokenized as (``\emph{ToCity}'', ``\emph{=}'', ``\emph{New}'', ``\emph{York}'').

\subsubsection{Local-level} For each slot-value pair $o^C_m$, we use a shared BLSTM to get local representations independently:
\begin{equation}
    (\mathbf{h}_{m1}^{l},\cdots,\mathbf{h}_{mT_m}^{l}) \leftarrow \text{BLSTM}_{\Theta_3}(\mathbf{c}_{m1},\cdots,\mathbf{c}_{mT_m})
\end{equation}
where $\mathbf{c}_{mj}$ is the embedding of $j$-th token\footnote{Every slot is also mapped to a trainable embedding vector.} in $o^C_m$. The local representation of $o^C_m$ is defined as $\mathbf{o}^C_m =\frac{1}{T_m}\sum_j\mathbf{h}_{mj}^{l}$, $\mathbf{o}^C_m \in \mathbb{R}^{2n}$.


\subsubsection{Global-level} Upon the local representations of all slot-value pairs in $o^C$, another BLSTM is applied to get global hidden features, $\mathbf{h}_m^g \in \mathbb{R}^{2n}$, $m\in\{1,\cdots,M\}$:
\begin{equation}
    (\mathbf{h}_1^{g},\cdots,\mathbf{h}_{M}^{g}) \leftarrow \text{BLSTM}_{\Theta_4}(\mathbf{o}^C_{1},\cdots,\mathbf{o}^C_{M}) .
\end{equation}


\textbf{Decoder:} In order to avoid generating redundant or missing slots in the prediction of sequence $\widetilde{x}$, the semantically controlled LSTM (SC-LSTM)~\cite{wen-etal-2015-semantically} is applied. The hidden vector at the $t$-th time step is computed by $(\mathbf{d}_t, \mathbf{s}_t)=\text{f}_\text{{SC-LSTM}}(\mathbf{\widetilde{x}}_{t-1}\oplus\mathbf{o}^I, (\mathbf{d}_{t-1}, \mathbf{s}_{t-1}))$, where $\mathbf{\widetilde{x}}_{t-1}$ is the embedding of the previously predicted token, $\mathbf{o}^I$ is the embedding of the given intent, and $\mathbf{d}_t$ is the hidden vector. Compared with LSTM, the SC-LSTM contains a slot-value state $\mathbf{s}_t$ which plays the role of sentence planning. $\mathbf{s}_t$ manipulates the slot-value features during the generation process in order to produce a hidden vector which accurately encodes the input semantics. The slot-value state is initialized with the original slots 1-hot vector $\mathbf{s}_0\in \mathbb{R}^{|\mathcal{S}|}$, where each element is zero except for the slots in $o^C$ and $\mathcal{S}$ is the set of all possible slots in the current domain, i.e.,
\begin{equation*}
    \mathbf{s}_{0i} = \left\{\begin{array}{ll}{1}, & {\text{if the } i\text{-th slot exists in } o^C;} \\ 
{0}, & {\text{otherwise.}}\end{array}\right.
\end{equation*}
Additional regularization term will be added to the final loss function for each sample ($\mathbf{s}_{|\widetilde{x}|}$ is the final slot state vector)
\begin{equation*}
    \mathcal{L}_{\text{SC}} = ||\mathbf{s}_{|\widetilde{x}|}||_2+\sum_{t=1}^{|\widetilde{x}|} \eta\xi^{||\mathbf{s}_t-\mathbf{s}_{t-1}||_2}
\end{equation*}
where $\eta=10^{-4},\xi=100$, $||\cdot||_2$ is $l_2$ norm. The first term is used to penalise generated sequences that failed to render all the required slots, while the second term discourages the
decoder from turning more than one slot off in a single time step.

The hidden vector is initialized by the aggregated encoding vectors, i.e. $\mathbf{d}_0 = \mathbf{W}_0\mathbf{z}_0^{\text{sem}}$, where $\mathbf{W}_0\in \mathbb{R}^{n\times2n}$, and $\mathbf{z}_0^{\text{sem}}\in \mathbb{R}^{2n}$ is an attention vector of the encoder hiddens, i.e.,
\begin{equation}
    \mathbf{z}_0^{\text{sem}}, (a_{01}^{\text{sem}},\cdots,a_{0M}^{\text{sem}}) \leftarrow \text{ATTN}_{\Theta_5}(\mathbf{o}^I, (\mathbf{h}_1^{g},\cdots,\mathbf{h}_{M}^{g})) .
\end{equation}


An output layer with the attention mechanism~\cite{luong2015effective} and the copying mechanism~\cite{see2017get} is applied on the SC-LSTM to predict tokens in $\widetilde{x}$. The attention weight for the current step $t$ of the decoder with the $m$-th slot-value pair in the encoder ($m\in\{1,\cdots,M\}$) and the attention vector are computed via
\begin{equation}
    \mathbf{z}_t^{\text{sem}}, (a_{t1}^{\text{sem}},\cdots,a_{tM}^{\text{sem}}) \leftarrow \text{ATTN}_{\Theta_6}(\mathbf{d}_t, (\mathbf{h}_1^{g},\cdots,\mathbf{h}_{M}^{g})) .
\end{equation}
Then we compute the vocabulary distribution 
\begin{equation}
    p_{\text{gen}}(\widetilde{x}_t|\widetilde{x}_{<t},y)=\text{softmax}_{\widetilde{x}_t}(\mathbf{W}_o(\mathbf{d}_t \oplus \mathbf{z}_t^{\text{sem}})) 
\end{equation}
where $\mathbf{W}_o \in \mathbb{R}^{|\mathcal{V}_{\widetilde{x}}|\times3n}$, and $|\mathcal{V}_{\widetilde{x}}|$ is the output vocabulary size. Generation terminates once an end-of-sequence token ``\texttt{EOS}'' is emitted.


Except for directly generation, the decoder also includes the copying mechanism to improve model generalization, which copies slots from the slot-value pairs in $o^C$. We use sigmoid gate function $\sigma$ to make a soft decision between generation and copy at each step $t$, i.e. $g_t = \sigma(\mathbf{v}_g^{\top}(\mathbf{d}_t \oplus \mathbf{z}_t^{\text{sem}}))$ and
\begin{equation*} 
p(\widetilde{x}_t|\widetilde{x}_{<t},y) = g_t p_{\text{gen}}(\widetilde{x}_t|\widetilde{x}_{<t},y) + (1-g_t)p_{\text{copy}}(\widetilde{x}_t|\widetilde{x}_{<t},y)
\end{equation*}
where $g_t\in[0,1]$ is the balance score, $\mathbf{v}_g$ is a weight vector. Distribution $p_{\text{copy}}(\cdot|\cdot)$ is defined over $M$ slots in $(o^C_1,\cdots,o^C_{M})$:
\begin{equation*}
    p_{\text{copy}}(\widetilde{x}_t|\widetilde{x}_{<t},y) = \left\{\begin{array}{ll}{
    a_{tm}^{\text{sem}},} & {\widetilde{x}_t \text{ is the slot of } o^C_m, m\in[1,M]} \\ {0,} & {\text {otherwise}}\end{array}\right.
\end{equation*} 

Afterward, we can get the final sentence $x$ by substituting each slot in $\widetilde{x}$ with the value in the corresponding slot-value pair. The loss function of SSG model given $x$ and $y$ is 
\begin{equation*}
    \mathcal{L}_{\text{SSG}}(y, x) = -\sum_{t=1}^{|\widetilde{x}|}\log p(\widetilde{x}_t|\widetilde{x}_{<t},y) + \mathcal{L}_{\text{SC}} . 
\end{equation*}

\section{Dual Semi-supervised NLU}
\label{sec:dual_semi_supervised_NLU}

\begin{algorithm}[t]
\caption{Dual Semi-supervised NLU.}
\label{alg:main}
\begin{algorithmic}[1]
\Require Labeled training set $\mathcal{D}_{\text{xy}}^L$; pure (unlabeled) sentences set $\mathcal{D}_{\text{x}}^U$; pure (unexpressed) semantic forms (i.e., intents and lists of slot-value pairs) set $\mathcal{D}_{\text{y}}^U$; beam search size $K$; weight factor $\delta$; maximum number of iterations $N$.


\State{Train sentence side language model $\text{LM}(\cdot)$ on data $\mathcal{D}_{\text{xy}}^L\cup\mathcal{D}_{\text{x}}^U$. Build lexicon database $\text{DB}(\cdot)$ and intent-slot co-occurence matrix $\text{COM}$ on $\mathcal{D}_{\text{xy}}^L\cup\mathcal{D}_{\text{y}}^U$.}

\State{Pre-train $\text{NLU}(\cdot|\Theta_{\text{NLU}})$ and $\text{SSG}(\cdot|\Theta_{\text{SSG}})$ models on $\mathcal{D}_{\text{xy}}^L$ by respectively minimizing the cross-entropy losses $\sum_{(x,y)\in\mathcal{D}_{\text{xy}}^L} \mathcal{L}_{\text{NLU}}(x, y)$ and $\sum_{(x,y)\in\mathcal{D}_{\text{xy}}^L}\mathcal{L}_{\text{SSG}}(y, x)$.}

\For{$i=1$ to $N$}
\Repeat

\State{Sample sentence $x\sim\mathcal{D}_{\text{xy}}^L\cup\mathcal{D}_{\text{x}}^U$}
\State{Sample semantic form $y\sim\mathcal{D}_{\text{xy}}^L\cup\mathcal{D}_{\text{y}}^U$}

\Comment{\textbf{Part 1: Dual Pseudo-Labeling Method}}
\State{Use the current NLU model to generate pseudo labels for $x$, i.e., $y' = \text{NLU}(x|\Theta_{\text{NLU}})$.}
\State{Use the current SSG model to generate pseudo sentences for $y$, i.e., $x' = \text{SSG}(y|\Theta_{\text{SSG}})$.}
\State{Update $\Theta_{\text{NLU}}$ and $\Theta_{\text{SSG}}$ on the generated pseudo-samples by minimizing $w_i ( \mathcal{L}_{\text{NLU}}(x, y') + \mathcal{L}_{\text{NLU}}(x', y) )$ and $w_i ( \mathcal{L}_{\text{SSG}}(y, x') + \mathcal{L}_{\text{SSG}}(y', x) )$ respectively.} 

\Comment{\textbf{Part 2: Dual Learning Method}}
\State{Produce $K$ semantic forms $y^{\prime 1},\cdots,y^{\prime K}$ using beam search according to
$\text{NLU}(x|\Theta_{\text{NLU}})$.}
\State{For each $y^{\prime k}$, generate $x^{\prime k}=\text{SSG}(y^{\prime k}|\Theta_{\text{SSG}})$.}
\State{For $k$-th sample, compute the total reward $r^k_1(x)$.}
\State{Compute gradients $\nabla_{\Theta_{\text{NLU}}}\psi_1(x)$ and $\nabla_{\Theta_{\text{SSG}}}\psi_1(x)$.}

\State{Produce $K$ sentences $x^{\prime 1},\cdots,x^{\prime K}$ using beam search according to
$\text{SSG}(y|\Theta_{\text{SSG}})$.}
\State{For each $x^{\prime k}$, generate $y^{\prime k}=\text{NLU}(x^{\prime k}|\Theta_{\text{NLU}})$.}
\State{For $k$-th sample, compute the total reward $r^k_2(y)$.}
\State{Compute gradients $\nabla_{\Theta_{\text{NLU}}}\psi_2(y)$ and $\nabla_{\Theta_{\text{SSG}}}\psi_2(y)$.}

\State{Update $\Theta_{\text{NLU}}$ with $\delta \nabla_{\Theta_{\text{NLU}}}\psi_1(x) + (1-\delta) \nabla_{\Theta_{\text{NLU}}}\psi_2(y)$; update $\Theta_{\text{SSG}}$ with $\delta \nabla_{\Theta_{\text{SSG}}}\psi_1(x) + (1-\delta) \nabla_{\Theta_{\text{SSG}}}\psi_2(y)$.}

\Comment{\textbf{Part 3: Raw Supervised Training}}
\State{Sample training pair $(x,y)\sim\mathcal{D}_{\text{xy}}^L$;}
\State{Update $\Theta_{\text{NLU}}$ and $\Theta_{\text{SSG}}$ by minimizing $\mathcal{L}_{\text{NLU}}(x, y)$ and $\mathcal{L}_{\text{SSG}}(y, x)$ respectively.}

\Until{all samples in $\mathcal{D}_{\text{xy}}^L, \mathcal{D}_{\text{x}}^U, \mathcal{D}_{\text{y}}^U$ are sampled.}

\EndFor
\end{algorithmic}
\end{algorithm}

In this section, we will describe our dual semi-supervised framework for the NLU task, which contains two methods: dual pseudo-labeling and dual learning. Algorithm \ref{alg:main} gives an overview of the dual semi-supervised NLU. Besides the labeled data $\mathcal{D}_{\text{xy}}^L$, there are two kinds of unlabeled data here: pure (unlabeled) sentences set $\mathcal{D}_{\text{x}}^U$, and pure (unexpressed) semantic forms (i.e., intents and lists of slot-value pairs without corresponding sentences) set $\mathcal{D}_{\text{y}}^U$.

\subsection{Dual Pseudo-Labeling Method}
\label{subsec:dual_pseudo_label}

Pseudo-Label are target labels for unannotated data as if they were true labels~\cite{lee2013pseudo}. We first pre-train the NLU and SSG models in a supervised fashion with labeled data. Given an unlabeled sentence $x$, we can use the NLU model to generate pseudo label $y'$. Symmetrically, we can also obtain $x'$ by utilizing the SSG model given an unexpressed semantic form $y$. In other words, we can create pseudo training samples $(x,y')$ and $(x',y)$ in addition to the existing labeled data, as shown in Algorithm \ref{alg:main} (part 1). 

Except for the supervised training stage with the labeled data, the NLU and SSG models can also be fine-tuned with the pseudo-samples by respectively minimizing losses $w_i ( \mathcal{L}_{\text{NLU}}(x, y') + \mathcal{L}_{\text{NLU}}(x', y) )$ and $w_i ( \mathcal{L}_{\text{SSG}}(y, x') + \mathcal{L}_{\text{SSG}}(y', x) )$ at the $i$-th iteration, where $w_i$ is an important coefficient. To prevent models stuck in poor local minima, we slowly increase $w_i$ such that greater confidence is assigned to pseudo-samples as training goes on. Concretely, $w_i=\frac{i}{N}$, where $N$ is the maximum number of iterations.


Besides unlabeled $x$ and unexpressed $y$, we also generate pseudo-samples starting from a sentence $x$ and a semantic form $y$ in the labeled data. Because it may rectify annotation noise and create various expressions for the same semantic form.

Previous work~\cite{lee2013pseudo,celikyilmaz2016new,xie2019self} related to the pseudo-labeling method only proposes to generate pseudo-labels $y'$ for unlabeled inputs $x$. To the best of our knowledge, we are the first to propose the dual pseudo-labeling method, which shares pseudo-samples between the NLU and SSG tasks and optimizes the NLU and SSG models iteratively.

\subsection{Dual Learning Method}
\label{subsec:dual_learning}

\begin{figure}[t] 
\centering
\includegraphics[width=0.48\textwidth]{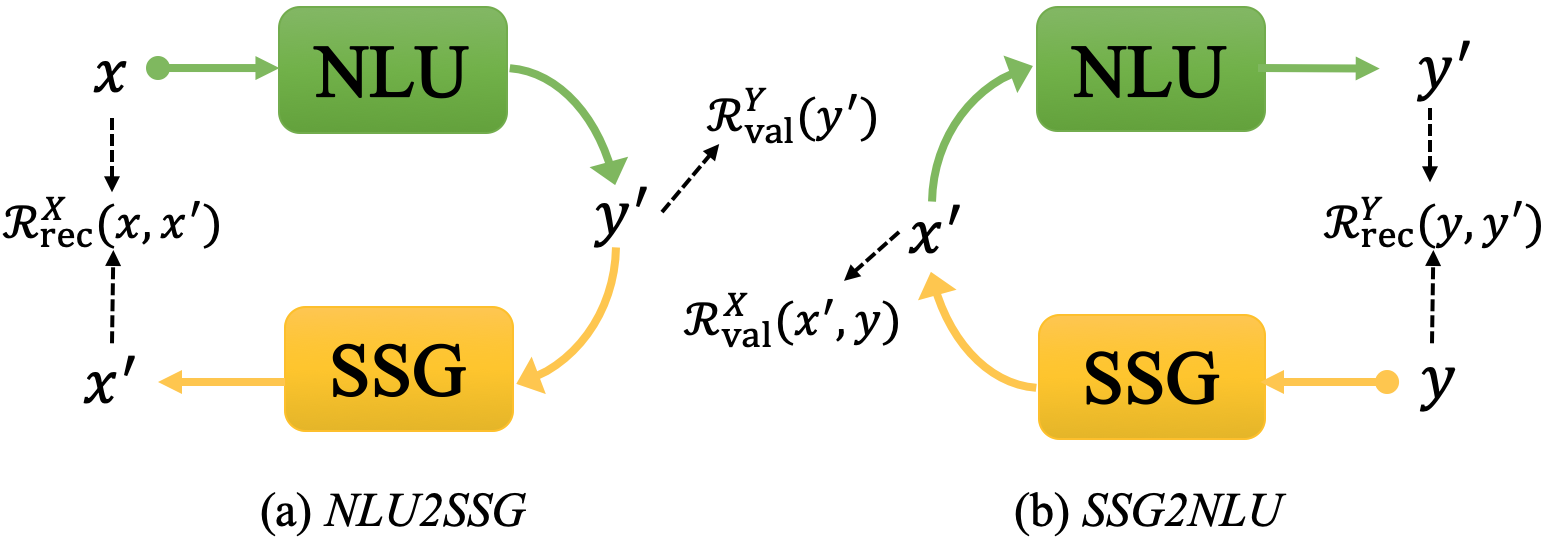}
\caption{An overview of dual learning method. The NLU and SSG models can form a closed cycle, which contains two directed loops \emph{NLU2SSG} and \emph{SSG2NLU} starting from a sentence $x$ and semantic form $y$ respectively.}
\label{fig:dual_learning}
\end{figure}

Besides the pseudo-labeling method where the NLU and SSG models are updated separately, we also propose to apply the dual learning method by jointly training the two models. We use one agent to represent the model of the primal task (NLU) and another agent to represent the model of the dual task (SSG). Then a two-agent game is designed in a closed loop which can provide quality feedback to the primal and dual models even if only sentences or semantic forms are available. As the feedback rewards are non-differentiable, reinforcement learning algorithm~\cite{sutton2018reinforcement} based on policy gradient~\cite{sutton2000policy} is applied for optimization.

As illustrated in Fig. \ref{fig:dual_learning}, two agents, NLU and SSG, participate in the collaborative game with two directed loops. 1) \emph{NLU2SSG} loop starts from a sentence, generates a possible semantic form by agent NLU and tries to reconstruct the original sentence by SSG. 2) \emph{SSG2NLU} loop starts from the opposite side. Each agent will obtain quality feedback depending on reward functions defined in the directed loops. The NLU and SSG models are pre-trained on the labeled data. Let $\Theta_{\text{NLU}}$ and $\Theta_{\text{SSG}}$ denote all the parameters of the NLU and SSG models respectively. A brief description of the dual learning algorithm is provided in Algorithm \ref{alg:main} (part 2), comprised of the two directed loops: 

\subsubsection{Loop NLU2SSG}
We sample a sentence $x$ from the union of labeled and unlabeled data randomly. Given $x$, the NLU model could produce $K$ possible semantic form $y^{\prime 1},\cdots,y^{\prime K}$ via beam search ($K$ is beam size). For each $y^{\prime k}$, we can obtain a validity reward $\mathcal{R}^Y_{\text{val}}(y^{\prime k})$ (a scalar) which reflects the likelihood of $y^{\prime k}$ being a valid semantic form. Afterwards, we pass $y^{\prime k}$ into the SSG model and get an output $x^{\prime k}$ by greedy decoding. Finally, we get a reconstruction reward $\mathcal{R}^X_{\text{rec}}(x, x^{\prime k})$ which forces the generated sentence $x^{\prime k}$ as similar to $x$ as possible. The rewards will be elucidated in Section \ref{sec:reward}. A coefficient $\alpha\in[0,1]$ is exploited to balance these two rewards in $r^k_1(x) = \alpha \mathcal{R}^Y_{\text{val}}(y^{\prime k}) + (1-\alpha) \mathcal{R}^X_{\text{rec}}(x, x^{\prime k})$.


By minimizing the negative expected reward $\psi_1(x)=\mathbb{E}[-\frac{1}{K}\sum_{k=1}^K r^k_1(x)]$ via policy gradient~\cite{sutton2000policy}, the stochastic gradients of $\Theta_{\text{NLU}}$ and $\Theta_{\text{SSG}}$ are computed as:
\allowdisplaybreaks
\begin{align*}
    \nabla_{\Theta_{\text{NLU}}}\psi_1(x) &= \frac{1}{K}\sum_{k=1}^K r^k_1(x) \nabla_{\Theta_{\text{NLU}}}\mathcal{L}_{\text{NLU}}(x,y^{\prime k})\\
    \nabla_{\Theta_{\text{SSG}}}\psi_1(x) &= \frac{1-\alpha}{K}\sum_{k=1}^K \mathcal{R}^X_{\text{rec}}(x, x^{\prime k}) \nabla_{\Theta_{\text{SSG}}}\mathcal{L}_{\text{SSG}}(y^{\prime k},x)
\end{align*}

\subsubsection{Loop SSG2NLU}
Symmetrically, we sample a semantic form $y$ from the labeled and unlabeled data randomly. Given $y$, the SSG model could generate $K$ possible sentences $x^{\prime 1},\cdots,x^{\prime K}$ via beam search. For each $x^{\prime k}$, we can obtain a validity reward $\mathcal{R}^X_{\text{val}}(x^{\prime k},y)$ which reflects whether the sampled natural language sentence $x^{\prime k}$ is well-formed and fluent. Afterwards, we feed $x^{\prime k}$ into the NLU model, and get the top-hypothesis $y^{\prime k}$. Finally, we get a reconstruction reward $\mathcal{R}^Y_{\text{rec}}(y, y^{\prime k})$ which forces $y^{\prime k}$ as similar to $y$ as possible. The rewards will be explained in Section \ref{sec:reward}. A coefficient $\beta\in[0,1]$ is exploited to balance these two rewards in $r^k_2(y) = \beta \mathcal{R}^X_{\text{val}}(x^{\prime k},y) + (1-\beta) \mathcal{R}^Y_{\text{rec}}(y, y^{\prime k})$.


By minimizing the negative expected reward $\psi_2(y)=\mathbb{E}[-\frac{1}{K}\sum_{k=1}^K r^k_2(y)]$ via policy gradient~\cite{sutton2000policy}, the stochastic gradients of $\Theta_{\text{SSG}}$ and $\Theta_{\text{NLU}}$ are computed as:
\begin{align*}
    \nabla_{\Theta_{\text{SSG}}}\psi_2(y) &= \frac{1}{K}\sum_{k=1}^K r^k_2(y) \nabla_{\Theta_{\text{SSG}}}\mathcal{L}_{\text{SSG}}(y,x^{\prime k})\\
    \nabla_{\Theta_{\text{NLU}}}\psi_2(y) &= \frac{1-\beta}{K}\sum_{k=1}^K \mathcal{R}^Y_{\text{rec}}(y, y^{\prime k}) \nabla_{\Theta_{\text{NLU}}}\mathcal{L}_{\text{NLU}}(x^{\prime k},y)
\end{align*}

To the best of our knowledge, we are the first to apply the dual learning algorithm to the intent detection and slot filling in NLU. Compared with the dual learning for neural machine translation~\cite{he2016dual} with only language models based reward, we introduce new validity and reconstruction rewards for structured data of NLU.

\subsubsection{Reward Design}
\label{sec:reward}

Here we will give some details about two validity and two reconstruction rewards introduced in the two directed loops above.

\textbf{Validity reward of} $\mathcal{R}^Y_{\text{val}}(y^{\prime})$ measures whether a possible semantic form $y^{\prime}$ is valid. It can be jointly evaluated on intents and slots from two relations:
\begin{itemize}
    \item \emph{Slot-value}: whether a given slot-value pair is valid, e.g. ``boston" is a valid city name for slot \texttt{FromCity}.
    \item \emph{Slot-intent}: whether a slot is likely to co-occur with respect to the predicted intent, e.g. \texttt{FromCity} often co-occurs with intent \texttt{find\_flight}.
\end{itemize}

To this end, a lexicon \textbf{d}ata\textbf{b}ase $\text{DB}(\cdot)$ is created from the training set, which specifies any possible value $v$ for each slot $s$. A \textbf{c}o-\textbf{o}ccurrence \textbf{m}atrix ($\text{COM}$) is leveraged from the training set, where $\text{COM}(i,s)\in \{0,1\}$ indicates whether the slot $s$ co-occurs with the user intent $i$. Concretely, $\mathcal{R}^Y_{\text{val}}(y^{\prime})$ is defined as
\begin{align*}
\text{score}(s, v) &= \underset{e\in \text{DB}(s)}{\text{max}} (1-\text{Edit\_Distance}(e,v)/|v|) \\
r_{\text{sv}}(o^{C\prime})&=\left\{\begin{matrix}
\frac{1}{|o^{C\prime}|}\sum_{(s,v)\in o^{C\prime}}\text{score}(s, v), & \text{if } |o^{C\prime}| \neq 0\\ 
1.0, &  \text{otherwise}
\end{matrix}\right. \\
r_{\text{si}}(o^{I\prime}, o^{C\prime})&=\left\{\begin{matrix}
\frac{1}{|o^{C\prime}|}\sum_{(s,v)\in o^{C\prime}}\text{COM}(o^{I\prime}, s), & \text{if } |o^{C\prime}| \neq 0\\ 
1.0, &  \text{otherwise}
\end{matrix}\right. \\
\mathcal{R}^Y_{\text{val}}(y^{\prime})&=\lambda\cdot r_{\text{sv}}(o^{C\prime})+(1-\lambda)\cdot r_{\text{si}}(o^{I\prime},o^{C\prime})
\end{align*}
where $y^{\prime}=(o^{I\prime},o^{C\prime })$ contains an intent $o^{I\prime}$ and a list of slot-value pairs $o^{C\prime}$, $\text{Edit\_Distance}(e,v)$ calculates a word-level edit distance between two values, and $\lambda$ is a weight factor.

\textbf{Validity reward of} $\mathcal{R}^X_{\text{val}}(x^{\prime},y)$ measures whether a generated natural language sentence $x^{\prime}$ is well-formed and fluent. We also evaluate it from two aspects:
\begin{itemize}
    \item \emph{Semantic integrity}: whether $x^{\prime}$ expresses all slots in the input $y$ precisely. This can be measured by a metric of slot accuracy, i.e. $\text{SlotAcc}(x^{\prime}, y)= 1-\frac{p+q}{m}$, where $m$ is the total number of slot-value pairs in $y$, $p$ and $q$ are the number of omitted and redundant slots in the delexicalized form of $x^{\prime}$ respectively.
    \item \emph{Word fluency}: the probability of $x^{\prime}$ to be a natural language sentence. We train a LSTM based language model~\cite{mikolov2010recurrent} with sentences of both the labeled and unlabeled data to evaluate the quality of $x^{\prime}$. Length-normalization~\cite{wu2016google} is applied to make a fair competition between short and long sentences, i.e. $\frac{1}{|x^{\prime}|} \log \text{LM}(x^{\prime})$.
\end{itemize}
A weight factor $\gamma$ is used to combine these two aspects:
\begin{equation*}
    \mathcal{R}^X_{\text{val}}(x^{\prime}, y)=\gamma\cdot \text{SlotAcc}(x^{\prime}, y)+(1-\gamma)\cdot \frac{1}{|x^{\prime}|} \log \text{LM}(x^{\prime}) .
\end{equation*}

\textbf{Reconstruction reward of} $\mathcal{R}^X_{\text{rec}}(x, x^{\prime})$ measures the similarity score between the finally generated sentence $x^{\prime}$ and the raw input $x$. The BLEU score~\cite{papineni2002bleu} is utilized:
\begin{equation*}
    \mathcal{R}^X_{\text{rec}}(x, x^{\prime}) = \text{BLEU}(x, x^{\prime}) .
\end{equation*}

\textbf{Reconstruction reward of} $\mathcal{R}^Y_{\text{rec}}(y, y^{\prime})$ reflects the similarity between the finally produced semantic form $y^{\prime}=(o^{I\prime}, o^{C\prime})$ and the raw input $y=(o^I, o^C)$. We use the slot-value $\text{F}_1$ score and intent accuracy to measure it, i.e.
\begin{equation*}
    \mathcal{R}^Y_{\text{rec}}(y, y^{\prime}) = \omega\mathbb{I}\{o^I=o^{I\prime}\} + (1-\omega)\text{F}_1(o^C, o^{C\prime}) .
\end{equation*}
where $\mathbb{I}$ is the indicator function and $\eta$ is  a weight factor.
\section{Experiments}
\label{sec:exp}

In this section, we first introduce the datasets and baselines with details of the experimental setup. Then, we compare the performance of our proposed methods with the baselines. Finally, extensive ablation studies are conducted for analysis. 

\subsection{Datasets}

We evaluate our proposed methods on two public datasets: Airline Travel Information Systems (ATIS) dataset~\cite{hemphill1990atis} and SNIPS Natural Language Understanding benchmark (SNIPS)~\cite{2018arXiv180510190C}. ATIS is a
widely used dataset in spoken language understanding, where audio recordings of people making flight reservations are collected and manually transcribed. SNIPS contains natural language corpus
collected in a crowdsourced fashion to benchmark
the performance of voice assistants. The statistical information on the two datasets are illustrated in Table \ref{tab:dataset}.


\begin{table}[]
  \caption{Dataset statistics.}
  \label{tab:dataset}%
  \centering{
    \begin{tabular}{c|cccc|cc}
    \hline
    Dataset  & Vocab Size & Train & Valid & Test & \#Slot & \#Intent  \\
    \hline \hline
    ATIS & 950 & 4478 & 500 & 893 & 83 & 18  \\
    \hline
    SNIPS & 14349 & 13084 & 700 & 700 & 39 & 7 \\
    \hline
    \end{tabular}
  }
\end{table}%


\subsection{Baselines}

We compare the proposed dual semi-supervised NLU with other alternatives:

\begin{itemize}
\item \emph{Supervised NLU} only exploits labeled data ($\mathcal{D}_{\text{xy}}^L$) for supervised learning, e.g. \emph{BLSTM-softmax}, \emph{BLSTM-CRF} and \emph{BLSTM-focus} methods described in Section \ref{sec:st_slu}.
\item \emph{Dual supervised learning}~\cite{su2019dual} incorporates the probabilistic duality of constraint (an additional regularization term) into the standard supervised learning.
\item \emph{Multi-task learning with unsupervised task} can exploit RNN-based language modelling ~\cite{rei2017semi,peters2017semi,oyl11-lan-icassp18,Aditya2018Unsupervised} and sequence-to-sequence based sentence auto-encoder~\cite{kim2017adversarial,zhu2018robust} to additionally utilize unlabeled sentences ($\mathcal{D}_{\text{x}}^U$). We implement the \emph{sentence auto-encoder} in our experiments.
\item The traditional \emph{pseudo-labeling (PL) method} without the dual task~\cite{tur2005combining,lee2013pseudo,celikyilmaz2016new} creates pseudo-samples for unlabeled sentences ($\mathcal{D}_{\text{x}}^U$) to perform data augmentation, using a pre-trained NLU model.
\item \emph{Template synthesis} method first extracts templates by converting each input sentence of the labeled data into its delexicalized form (e.g. ``\emph{show me flights from} $\langle\texttt{FromCity}\rangle$ \emph{to} $\langle \texttt{ToCity} \rangle$''). Afterwards, we synthesize additional labeled samples for the supervised training by replacing slot types in each template with the corresponding values provided in unexpressed semantic forms ($\mathcal{D}_{\text{y}}^U$).
\end{itemize}

\begin{table*}[t]
  \caption{Slot $\text{F}_1$ scores and intent accuracies of different methods on ATIS and SNIPS datasets. We randomly select \{5,10,20,30,50\}\% of the training set as labeled data, and leave the rest to simulate unlabeled data. The results in bold black are the best slot F$_1$ scores and intent accuracies. $^\ddagger$ indicates our results that significantly outperform the best baseline.}
  \label{tab:main_results_on_atis_snips}%
  \centering{
    \begin{tabular}{l|l|p{0.6cm}p{0.6cm}p{0.6cm}p{0.6cm}p{0.6cm}p{0.6cm}|p{0.6cm}p{0.6cm}p{0.6cm}p{0.6cm}p{0.6cm}p{0.6cm}}
    \hline
    \multicolumn{2}{l|}{\multirow{2}*{\diagbox{\textbf{Method}}{\textbf{\ \ \ \ Slot $\text{F}_1$} (\%)}}} & \multicolumn{6}{c|}{\textbf{ATIS}} & \multicolumn{6}{c}{\textbf{SNIPS}} \\
    \cline{3-14}
    \multicolumn{2}{c|}{} & 5\% & 10\% & 20\% & 30\% & 50\% & 100\% & 5\% & 10\% & 20\% & 30\% & 50\% & 100\% \\
    \hline
    \multirow{2}*{supervised} & \textbf{BLSTM-focus} (backbone) & 82.92 & 89.48 & 92.66 & 93.54 & 95.45 & 95.79 & 87.89 & 91.23 & 93.54 & 94.45 & 94.92 & \textbf{96.44} \\
    & \ \ + dual supervised learning & 83.88 & 89.37 & 93.21 & 94.30 & 95.40 & 96.03 & 88.45 & 91.13 & 93.77 & 94.45 & 94.89 & 96.06 \\
    \hline
    \multirow{3}*{\begin{tabular}[c]{@{}l@{}}semi-\\ supervised\end{tabular}} & \ \ + sentence auto-encoder & 83.16 & 89.65 & 92.74 & 94.52 & 95.36 & 95.87 & 87.83 & 90.29 & 93.20 & 94.68 & 94.78 & 95.89 \\ 
    & \ \ + pseudo-labeling (PL) & 84.75 & 90.08 & 94.07 & 94.91 & 95.52 & 95.75 & 90.67 & 91.89 & 93.89 & 94.29 & 95.06 & 96.00 \\
    & \ \ + template synthesis & 86.10 & 90.62 & 94.35 & 94.94 & 95.27 & - & 90.40 & 92.94 & 93.94 & 94.22 & 94.37 & - \\
    \hline
    \multirow{3}*{\begin{tabular}[c]{@{}l@{}}+ dual task\\(ours)\end{tabular}} & \ \ + dual PL & \textbf{89.58}$^\ddagger$ & 93.49$^\ddagger$ &  94.88$^\ddagger$ & \textbf{95.90}$^\ddagger$ & 96.02$^\ddagger$ & 95.82 & 93.86$^\ddagger$ & 94.46$^\ddagger$ & \textbf{95.53}$^\ddagger$ & 95.23$^\ddagger$ & 95.29 & 96.22 \\
    & \ \ + dual learning &  88.92$^\ddagger$ & 93.40$^\ddagger$ & 95.09$^\ddagger$ & 95.50$^\ddagger$ & 95.70 & 96.00 & 93.85$^\ddagger$ & 94.18$^\ddagger$ & 95.31$^\ddagger$ & 95.08$^\ddagger$ & \textbf{95.45}$^\ddagger$ & 95.86 \\
    & \ \ + dual PL + dual learning &  \textbf{89.58}$^\ddagger$ & \textbf{93.53}$^\ddagger$ & \textbf{95.37}$^\ddagger$ & 95.85$^\ddagger$ & \textbf{96.14}$^\ddagger$ & \textbf{96.37}$^\ddagger$ & \textbf{94.00}$^\ddagger$ & \textbf{94.51}$^\ddagger$ & 95.22$^\ddagger$ & \textbf{95.34}$^\ddagger$ & 95.25 & 96.11 \\
    \hline
        \hline
    \multicolumn{2}{l|}{\multirow{2}*{\diagbox{\textbf{Method}}{\textbf{Intent Acc} (\%)}}} & \multicolumn{6}{c|}{\textbf{ATIS}} & \multicolumn{6}{c}{\textbf{SNIPS}} \\
    \cline{3-14}
    \multicolumn{2}{c|}{} & 5\% & 10\% & 20\% & 30\% & 50\% & 100\% & 5\% & 10\% & 20\% & 30\% & 50\% & 100\% \\
    \hline
    \multirow{2}*{supervised} & \textbf{BLSTM-focus} (backbone) & 89.03 & 92.61 & 94.40 & 94.85 & \textbf{98.54} & 98.43 & 97.86 & 98.14 & 98.00 & 98.29 & 98.71 & \textbf{99.14} \\
    & \ \ + dual supervised learning & 89.36 & 92.16 & 94.96 & 95.41 & 98.32 & 98.54 & 97.00 & 98.14 & 98.14 & 98.14 & 99.00 & \textbf{99.14} \\
    \hline
    \multirow{3}*{\begin{tabular}[c]{@{}l@{}}semi-\\ supervised\end{tabular}} & \ \ + sentence auto-encoder & 88.80 & 92.50 & 95.18 & 94.62 & 98.32 & 98.32 & 97.57 & 97.86 & 97.86 & 98.00 & 98.71 & 98.86 \\
    & \ \ + pseudo-labeling (PL) & 89.47 & 92.50 & 95.18 & 94.85 & 98.32 & 98.32 & 97.57 & 98.00 & 98.14 & 98.00 & 99.00 & \textbf{99.14} \\
    & \ \ + template synthesis & 90.05 & 92.05 & 94.06 & 94.51 & 98.10 & - & 97.86 & 98.00 & \textbf{98.57} & 98.14 & 98.86 & - \\
    \hline
    \multirow{3}*{\begin{tabular}[c]{@{}l@{}}+ dual task\\(ours)\end{tabular}} & \ \ + dual PL & 90.37 & 93.28 & 94.62 & \textbf{96.08}$^\ddagger$ & 98.43 & \textbf{98.66} & \textbf{98.57}$^\ddagger$ & 98.14 & 98.43 & 98.43 & 98.71 & \textbf{99.14} \\
    & \ \ + dual learning & 89.81 & 93.28 & \textbf{95.30} & 95.86 & \textbf{98.54} & 98.54 & 98.29 & 98.14 & 98.29 & \textbf{98.57} & 98.57 & 98.86 \\
    & \ \ + dual PL + dual learning &  \textbf{90.48} & \textbf{93.51}$^\ddagger$ & 95.18 & 95.30 & \textbf{98.54} & 98.54 & 98.29 & \textbf{98.43} & \textbf{98.57} & 98.14 & \textbf{99.14} & 98.86 \\
    \hline
    \end{tabular}%
  }
\end{table*}%

\subsection{Experimental Setup}
\label{sec:exp-setup}

\subsubsection{Training Details}

The word embeddings with 400 dimensions are initialized by concatenating pre-trained Glove embeddings\footnote{\url{http://nlp.stanford.edu/data/glove.840B.300d.zip}}~\cite{pennington2014glove} and character embeddings~\cite{hashimoto-etal-2017-joint}, which can be updated during training. The hidden size $n$ is 256. Hyper-parameters $\alpha, \beta, \gamma, \omega, \delta$ are set to $0.5$, and $\lambda$ is $0.25$ empirically. For the dual learning, the beam size $K$ is set to $5$. The network parameters are randomly initialized under the uniform distribution $[-0.2, 0.2]$, except for the pre-trained word embeddings. We use optimizer Adam \cite{kingma2014adam} with learning rate $0.001$ for all experiments. The \emph{dropout} with a probability of $0.5$ is applied to the non-recurrent connections during the training stage. The batch size is $16$ for all datasets. The maximum norm for gradient clipping is set to 5, and we use $l_2$ norm regularization on all weights with factor $1e\text{-}5$ to avoid over-fitting. We keep the learning rate for $50$ epochs and save the parameters that give the best performance on the validation set. Finally, we report the intent accuracy and $\text{F}_1$-score of slot-value pairs on the test set with parameters that have achieved the best average of intent accuracy and slot $\text{F}_1$-score on the validation set. The $\text{F}_1$-score is calculated using CoNLL evaluation script\footnote{\url{https://www.clips.uantwerpen.be/conll2000/chunking/output.html}}.

Besides using pre-trained word embeddings, some advanced pre-trained language models (e.g., ELMo~\cite{peters-etal-2018-deep}, BERT~\cite{devlin2018bert}) can also be used to get input embeddings. It is investigated in the following ablation studies. We employ the pre-trained BERT model (\emph{bert-base-cased}) with 12 layers of 768 hidden units and 12 self-attention heads\footnote{\url{https://github.com/google-research/bert}}. We update all the parameters using the Adam with a learning rate $5e\text{-}5$.


\subsubsection{Data settings for semi-supervised learning}
\label{sec:data_setting_for_semi_supervised_learning}

To evaluate the effectiveness and efficiency of different methods for semi-supervised NLU, we discuss the experimental configuration for semi-supervised settings below. In order to simulate the annotation scarcity problem in the real world, a part of the training set is kept as fully labeled data ($\mathcal{D}_{\text{xy}}^L$), and the rest is left as unpaired sentences and semantic forms ($\mathcal{D}_{\text{x}}^U$ and $\mathcal{D}_{\text{y}}^U$ respectively) to simulate unlabeled data. For the part of labeled data, we randomly select $5, 10, 15, 20, 30$ and $50$ percent of the training set in each dataset for experiments.


\subsubsection{Significance Test} We use McNemar's test to establish the statistical significance of a method over another ($\text{p}<0.05$).

\subsection{Overall Results}
\label{sec:exp-performance}

\begin{table}[t]
  \caption{Comparison among BLSTM-softmax, BLSTM-CRF and BLSTM-Focus for supervised NLU on ATIS and SNIPS datasets.}
  \label{tab:comparison_of_blstm_crf_focus}%
  \centering{
    \begin{tabular}{|l|cc|cc|}
    \hline
    \multirow{2}*{\textbf{Method}} & \multicolumn{2}{c|}{\textbf{ATIS}} & \multicolumn{2}{c|}{\textbf{SNIPS}} \\
    \cline{2-5}
    & Slot $\text{F}_1$ & Intent Acc & Slot $\text{F}_1$ & Intent Acc  \\
    \hline
    BLSTM-softmax & 95.50 & 98.21 & 94.96 & 98.86 \\
    BLSTM-CRF & 95.62 & 98.32 & 96.34 & 98.86\\
    BLSTM-focus & \textbf{95.79} & \textbf{98.43}  & \textbf{96.44} & \textbf{99.14} \\
    \hline
    \end{tabular}%
  }
\end{table}%

We first compare different methods on ATIS and SNIPS datasets with the simulated semi-supervised settings as well as full training samples, using the pre-trained word embeddings. Table \ref{tab:main_results_on_atis_snips} shows slot $\text{F}_1$ scores and intent accuracies of baselines and our methods on ATIS and SNIPS, then we can find that:
\begin{enumerate}
    \item Intent detection is a much easier sub-task than slot filling. We can see that the performance gap of intent accuracy between using 5\% and 100\% labeled data is lower than that of slot $\text{F}_1$ score, especially on the SNIPS dataset. Meanwhile, there is little difference among various methods with respect to intent accuracy, whereas our methods can achieve the best in most cases. 
    
    \item For \emph{supervised} NLU, we choose BLSTM-focus as our backbone model of NLU, rather than BLSTM-softmax and BLSTM-CRF. As shown in Table \ref{tab:comparison_of_blstm_crf_focus}, the BLSTM-focus model can achieve the best performance on ATIS and SNIPS with full training data.
    
    
    \item Three baselines of \emph{semi-supervised} learning NLU can improve performances by exploiting unlabeled data in most cases, where ``\emph{+ sentence auto-encoder}'' adds a sequence-to-sequence based sentence reconstruction task, and ``\emph{+ pseudo-labeling (PL)}'' uses the existing NLU model to generate pseudo-labels for unannotated sentences. ``\emph{+ template synthesis}'' exploits unexpressed semantic forms and the labeled data to synthesize more labeled samples for training.
    \item Compared with the traditional pseudo-labeling method (\emph{+ PL}) without the dual task, our proposed dual pseudo-labeling method (\emph{+ dual PL}) can get improvements by taking advantage of unexpressed semantic forms.
    \item The proposed dual learning-based method can also make improvements over the baselines. Different from the dual pseudo-labeling method, it involves validity reward and reconstruction reward to estimate (soft) importances of generated sentences or semantic forms.
    \item Finally, we combine the two proposed methods (as shown in Algorithm \ref{alg:main}) and obtain further improvements. In most cases, the combination (\emph{+ dual PL + dual learning}) can obtain the best performances especially on slot $\text{F}_1$ scores.

\begin{table}[]
  \caption{Data analysis of test sets compared with training sets.}
  \label{tab:data_analysis_of_test_sets}%
  \centering{
    \begin{tabular}{|c|c|c|}
    \hline
    Dataset  & \#Unseen delexicalized form & \#Unseen slot-value pairs  \\
    \hline 
    ATIS & 680 & 169  \\
    \hline
    SNIPS & 421 & 522 \\
    \hline
    \end{tabular}
  }
\end{table}%

    \item Our methods can even get improvements with 100\% labeled data (i.e. no unlabeled data), e.g., we get $96.37\%$ slot $\text{F}_1$ score on ATIS. However, our methods do not outperform the purely supervised method with 100\% labeled data (slot $\text{F}_1$ is $96.44\%$) on SNIPS. The reason may be that the test set of ATIS contains more unseen delexicalized forms, while the test set of SNIPS includes more unseen slot-value pairs, as shown in Table \ref{tab:data_analysis_of_test_sets}. Our methods applied to fully labeled data are likely to generate varied natural language expressions (sentences) for existing semantic forms. Therefore, our methods fail to get improvements on SNIPS due to lots of unseen slot-value pairs.
\end{enumerate}

\subsection{Analysis}
\label{sec:ablation}

In Section \ref{sec:exp-performance}, significant improvements of two metrics have been witnessed on the two datasets. However, we would like to figure out the potential factors for the improvement. In this sub-section, we will show ablation studies on the SSG model, the dual pseudo-labeling and dual learning methods to reveal the effects of different components. Finally, we analyze the effect of BERT in our framework. 

\subsubsection{Ablation studies of the SSG model}
To verify the effectiveness of the SSG model for the dual task of NLU, we apply ablation studies of supervised SSG on ATIS and SNIPS datasets, as shown in Table \ref{tab:ablation_of_SSG}. BLEU score~\cite{papineni2002bleu} is exploited to measure the similarity between generated sentences and references in the test set. We also utilize the slot accuracy mentioned in Section \ref{sec:reward} to measure the semantic integrity of generated sentences. From the result of ``\emph{(-) w/o feeding intent}'' row, we can observe that the intent is essential for BLEU scores, whereas intent detection is much easy in NLU. The \emph{global-level BLSTM} in the encoder, \emph{copy mechanism} and \emph{SC-LSTM cell} in the decoder are also important components of the SSG model.

Besides the supervised training, we may further want to know whether the SSG model will be improved in our proposed dual semi-supervised NLU. As shown in Table \ref{tab:ablation_of_SSG_with_dual}, the \emph{dual pseudo-labeling} and \emph{dual learning} methods can also improve the performance of the SSG model as well as the NLU model, where 10\% and 5\% of the training sets are selected as labeled data in ATIS and SNIPS respectively.

\begin{table}[t]
  \caption{Evaluations of the SSG model for the dual task of NLU, which is supervised by full training sets on ATIS and SNIPS respectively.}
  \label{tab:ablation_of_SSG}%
  \centering{
    \begin{tabular}{l||cc|cc}
    \hline
    \multirow{2}*{\textbf{Model}} & \multicolumn{2}{c|}{\textbf{ATIS}} & \multicolumn{2}{c}{\textbf{SNIPS}} \\
    \cline{2-5}
    & BLEU & Slot Acc & BLEU & Slot Acc \\
    \hline \hline
    supervised SSG  & 47.17 & 97.72 & 39.18 & 100.00 \\
    \hline
    \quad(-) w/o feeding intent &  44.86 & 98.26 & 38.15 & 99.70 \\
    \quad(-) w/o global BLSTM & 41.14 & 96.15 & 31.65 & 98.95 \\
    \quad(-) w/o copy mechanism & 44.08 & 96.58 & 38.67 & 99.98 \\
    \quad(-) w/o SC-LSTM &  46.78 & 97.25 & 37.99 & 100.00 \\
    \hline
    \end{tabular}%
  }
\end{table}%

\begin{table}[t]
  \caption{Evaluations of the SSG model in the proposed dual semi-supervised NLU.}
  \label{tab:ablation_of_SSG_with_dual}%
  \centering{
    \begin{tabular}{l|cc|cc}
    \hline
    \multirow{2}*{\textbf{Method}} & \multicolumn{2}{c|}{\textbf{ATIS} (10\%)} & \multicolumn{2}{c}{\textbf{SNIPS} (5\%)} \\
    \cline{2-5}
    & BLEU & Slot Acc & BLEU & Slot Acc  \\
    \hline
    supervised SSG & 39.53 & 87.49 & 29.94 & 94.85 \\
    \hline
    \ \ + dual PL & 40.28 & 93.90 & 35.26 & 98.85 \\
    \ \ + dual learning & 38.84 & 91.30 & 32.19 & 98.10 \\
    \ \ + dual PL + dual learning & 41.85 & 95.10 & 36.61 & 99.43 \\
    \hline
    \end{tabular}%
  }
\end{table}%


\subsubsection{Ablation studies of the dual pseudo-labeling method} 
The dual pseudo-labeling method creates pseudo-samples in two ways: a) obtaining predicted semantic forms of sentences with the NLU model and b) generating sentences with the SSG model given intents and slot-value pairs. Experiments are conducted to make a comparison of these two ways, as shown in Table \ref{tab:ablation_of_dual_pseudo_labeling}. From the results, we can find that \emph{pseudo-samples created from SSG} are more essential. The reason may be that the SSG model tends to generate sentences semantically consistent with the given semantic forms though the sentences are not natural enough. However, the NLU model may predict wrong semantic labels. Meanwhile, without \emph{pseudo-samples from SSG model}, the dual pseudo-labeling method reduces to the traditional pseudo-labeling without dual task.

From the result of ``\emph{(+) $w_i\text{=}1$}'' row in Table \ref{tab:ablation_of_dual_pseudo_labeling}, we can see that the increasing coefficient ($w_i$) at each iteration helps. We believe that the NLU and SSG models updated with more iterations could provide more qualified pseudo-samples. Therefore, the performance decreases if we keep using pseudo-samples generated at the first iteration, as shown in ``\emph{(-) w/o iterative generation}'' row.

\begin{table}[t]
  \caption{Ablation studies of the dual pseudo-labeling method.}
  \label{tab:ablation_of_dual_pseudo_labeling}%
  \centering{
    \begin{tabular}{|l|cc|}
    \hline
    \multirow{2}*{\textbf{Method}} & \multicolumn{2}{c|}{\textbf{SNIPS} (5\%)} \\
    \cline{2-3}
    & Slot $\text{F}_1$ & Intent Acc  \\
    \hline \hline
    + dual PL & 93.86 & 98.57 \\
    \hline
    \quad (-) w/o pseudo-samples from NLU model & 93.51 & 98.00 \\
    \quad (-) w/o pseudo-samples from SSG model & 90.67 & 97.57 \\
    \quad (+) $w_i=1$ & 93.65 & 98.14 \\
    \quad (-) w/o iterative generation & 91.49 & 97.71 \\
    \hline
    \end{tabular}%
  }
\end{table}%


\subsubsection{Ablation studies of the dual learning method}
Several experiments are conducted to show the effects of different components in the dual learning method, as illustrated in Table \ref{tab:ablation_of_dual_learning}. From the results of ``\emph{(-) w/o unlabeled sentences}'' and ``\emph{(-) w/o unexpressed semantic forms}'' rows, we can find that unexpressed semantic forms are more important, which is consistent with the findings in the ablation studies of the dual pseudo-labeling method. It may facilitate semi-supervised NLU, since semantic forms are well-structured and could be easily synthesized under domain knowledge. The last two rows show that two kinds of rewards are essential, while the validity rewards impact more on the slot $\text{F}_1$ score.

\begin{table}[t]
  \caption{Ablation studies of the dual learning method.}
  \label{tab:ablation_of_dual_learning}%
  \centering{
    \begin{tabular}{|l|cc|}
    \hline
    \multirow{2}*{\textbf{Method}} & \multicolumn{2}{c|}{\textbf{SNIPS} (5\%)} \\
    \cline{2-3}
    & Slot $\text{F}_1$ & Intent Acc  \\
    \hline
    \hline
    + dual learning & 93.85 & 98.29 \\
    \hline
    \quad (-) w/o unlabeled sentences & 92.96 & 98.14 \\
    \quad (-) w/o unexpressed semantic forms & 91.41  & 97.71 \\ 
    \quad (-) w/o validity rewards & 91.55 & 97.71 \\
    \quad (-) w/o reconstruction rewards & 93.74 & 97.86 \\
    \hline
    \end{tabular}%
  }
\end{table}%

\subsubsection{Effect of BERT}
Besides using pre-trained word embeddings, the BERT model can also be used to get input embeddings\footnote{We only consider BERT embeddings of the first subword if a word is broken into multiple subwords.}. However, it is orthogonal to the investigation of semi-supervised NLU. Table \ref{tab:effect_of_bert} shows results on ATIS and SNIPS, where 10\% and 5\% of the training sets are selected as labeled data in respective. The results show that a pre-trained BERT model can further enhance our dual semi-supervised NLU as well as the baseline. Although BERT embeddings can bridge the gap between our method and the baseline, the dual semi-supervised NLU still outperforms the baseline significantly.

\begin{table}[t]
  \caption{Slot $\text{F}_1$ scores and intent accuracies of BERT-based models on the two datasets.}
  \label{tab:effect_of_bert}%
  \centering{
    \begin{tabular}{|l|c|cc|cc|}
    \hline
    \multirow{2}*{\textbf{Method}} & \multirow{2}*{\begin{tabular}[c]{@{}c@{}}with\\ BERT\end{tabular}} & \multicolumn{2}{c|}{\textbf{ATIS} (10\%)} & \multicolumn{2}{c|}{\textbf{SNIPS} (5\%)} \\
    \cline{3-6}
    & & Slot & Intent & Slot & Intent \\
    \hline
    \text{BLSTM-focus} & \xmark &  89.48 & 92.61 & 87.89 & 97.86 \\
    + dual PL + dual learning & \xmark & 93.53 & 93.51 & 94.00 & 98.29  \\
    \hline
    \text{BLSTM-focus} & \cmark & 91.41 & 93.51 & 91.53 & 98.14 \\
    + dual PL + dual learning & \cmark & 94.14 & 94.29 & 95.58 & 98.43 \\
    \hline
    \end{tabular}%
  }
\end{table}%


\subsection{Compared with the Previous Results of the Supervised NLU}
\label{sec:sota}

Finally, we make a comparison with the previous results on ATIS and SNIPS datasets using full training sets, as illustrated in Table \ref{tab:sota}. Our proposed method (\emph{+ dual PL + dual learning}) can achieve the state-of-the-art performances on the two datasets, but not significantly outperforming the previous state-of-the-art. We can also find that BERT boosts the performance of ATIS less than SNIPS, which may occur due to the much smaller vocabulary of ATIS. Our method enhanced with BERT gets a decrease of slot $\text{F}_1$ score (from $96.4\%$ to $96.0\%$) and an increase of intent accuracy (from $98.5\%$ to $99.1\%$) on ATIS, while it achieves a better average score. It shows that our proposed method can also work well in fully supervised settings.

Moreover, our proposed method modeling the two sub-tasks (intent detection and slot filling) independently can surpass several approaches which consider dependence between these two sub-tasks ~\cite{goo-etal-2018-slot,qin-etal-2019-stack,li2018self,haihong2019novel}. We believe that it will be valuable to investigate methods of jointly modeling the intent detection and slot filling in our proposed framework. We leave it as a future work to explore the backbone NLU model.



\begin{table}[t]
  \centering{
\begin{threeparttable}[b]
  \caption{Comparison with previous results of NLU on ATIS and SNIPS.}
  \label{tab:sota}%
    \begin{tabular}{l|l|cc|cc}
    \hline
    \multicolumn{2}{c|}{\multirow{2}*{\textbf{Method}}} & \multicolumn{2}{c|}{\textbf{ATIS}} & \multicolumn{2}{c}{\textbf{SNIPS}} \\
    \cline{3-6}
    \multicolumn{2}{l|}{} & Slot & Intent & Slot & Intent  \\
    \hline
    \multirow{9}*{\begin{tabular}[c]{@{}c@{}}w/o\\ BERT\end{tabular}} & Joint Seq.~\cite{hakanni-tur2016multidomain}\tnote{$*$} & 94.3 & 92.6 &  87.3 & 96.9 \\
     & Attention BiRNN~\cite{liu2016attention}\tnote{$*$} & 94.2 & 91.1 & 87.8 & 96.7 \\
     & Slot-Gated~\cite{goo-etal-2018-slot} & 95.2 & 94.1 & 88.8 & 97.0 \\
     & Self-Attentive Model~\cite{li2018self}\tnote{$*$}  & 95.1 & 96.8 & 90.0 & 97.5\\
     & Bi-Model~\cite{wang2018bi}\tnote{$*$}  & 95.5 & 96.4 & 93.5 & 97.2\\
     & CAPSULE-NLU~\cite{zhang2019joint} & 95.2 & 95.0 & 91.8 & 97.3\\
     & ELMo-Light for SLU~\cite{Aditya2018Unsupervised} & 95.4 & 97.3 & 93.3 & 98.8 \\
     & SF-ID Network~\cite{haihong2019novel} & 95.8 & 97.1 & 92.2 & 97.3\\
     & Stack-Propagation~\cite{qin-etal-2019-stack}  & 95.9 & 96.9 & 94.2 & 98.0\\
     & \textbf{our method} & \textbf{96.4} & {98.5} & {96.1} &  {98.9}\\
    \hline
     \multirow{4}*{\begin{tabular}[c]{@{}c@{}}w/\\ BERT\end{tabular}} & Multi-ling. Joint BERT~\cite{castellucci2019multi} & 95.7 & 97.8 & 96.2 & 99.0 \\
     & Joint BERT SLU~\cite{chen2019bert}  & {96.1} & 97.5 &  97.0 & 98.6\\
     & Stack-Prop. + BERT~\cite{qin-etal-2019-stack} & {96.1} & 97.5 & 97.0 & 99.0\\
     & \textbf{our method + BERT} & 96.0 & \textbf{99.1} & \textbf{97.1} & \textbf{99.1} \\
    \hline
    \end{tabular}%
    \begin{tablenotes}
    \item [$*$] indicates a result borrowed from Qin et al.~\cite{qin-etal-2019-stack}.
    \end{tablenotes}
\end{threeparttable}
  }
\end{table}%


\subsection{Applicability to Other NLU Tasks}
\label{sec:other_nlu}

Except for the intent detection and slot filling tasks focused in this paper, there are other NLU tasks producing structured predictions, like knowledge base question answering (semantic parsing), semantic role labeling, etc. Thus, in theory, the proposed semi-supervised NLU framework could be applied in other NLU tasks by adjusting architectures of the primal and dual models to the new NLU task. Furthermore, we should redefine reward functions to suit different NLU tasks. 

Here we give an example of applying the proposed semi-supervised NLU framework into the semantic parsing task~\cite{berant-liang-2014-semantic}. 
Following Jia and Liang~\cite{jia-liang-2016-data}, we formulate the primal and dual tasks of semantic parsing as sequence generation problems, and adopt sequence-to-sequence RNN models (with an attention mechanism~\cite{bahdanau2014neural} or a pointer network~\cite{see2017get}) for the primal and dual tasks. For validity rewards, language models in sentence side and logical form side are exploited. For reconstruction rewards, we adopt log probabilities of recovered sequences~\cite{he2016dual}. Our method is evaluated in two benchmarks of semantic parsing: ATIS~\cite{dong2018coarse} and \textsc{Overnight}~\cite{wang2015building}. To simulate the semi-supervised setting, we randomly select 50\% training samples as fully labeled data (Section \ref{sec:data_setting_for_semi_supervised_learning}). The results are shown in Table \ref{tab:exp_on_semantic_parsing}, where two backbone models (``Attention'', ``Attention + Pointer Net.'') are provided. 
We can find that our method can obtain improvements in both two benchmarks for different backbone models.

\begin{table}[]
  \caption{Test accuracies on ATIS and Overnight in semi-supervised settings (the ratio of fully labeled data is 50\%).}
  \label{tab:exp_on_semantic_parsing}%
  \centering{
    \begin{tabular}{|l|c|c|}
    \hline
    \textbf{Method}  & ATIS & \textsc{Overnight}  \\
    \hline 
    Attention & 78.6 & 65.4  \\
    \quad + dual learning & 80.6 & 71.5  \\
    \hline
    Attention + Pointer Net. & 84.8 & 65.2 \\
    \quad + dual learning & 86.2 & 71.4  \\
    \hline
    \end{tabular}
  }
\end{table}%

\section{Conclusion}
\label{sec:conclusion}

This paper has introduced a dual task for SLU, which is semantic-to-sentence generation (SSG). It is incorporated in our proposed dual semi-supervised NLU to utilize unexpressed semantic forms as well as unlabeled sentences. The dual semi-supervised NLU includes the dual pseudo-labeling and dual learning methods which can learn the NLU and SSG models iteratively in the closed-loop of the primal and dual tasks. The proposed approaches are evaluated on two public datasets (ATIS and SNIPS). From the experimental results, we find that the dual semi-supervised NLU involving SSG could significantly improve the performances over traditional semi-supervised methods. We also provide extensive ablation studies to verify the effectiveness of our methods. Meanwhile, our methods can also achieve the state-of-the-art performance on the two datasets in the supervised setting.

The proposed framework of dual semi-supervised NLU shows promising perspectives of future improvements.
\begin{itemize}
    \item Exploiting semantic forms for semi-supervised learning could be more affordable and effective than collecting in-domain sentences, since semantic forms are well-structured and could be automatically synthesized under domain knowledge.
    \item Although our proposed framework is agnostic of the backbone model of NLU task, it will be meaningful to explore different NLU models and showcase the bottleneck where the dual learning hits its performance limit.
    \item Validity and reconstruction rewards are important for softly validating pseudo-samples. We will explore appropriate rewards to improve the effectiveness and efficiency of the dual semi-supervised NLU in our future work.
    \item This work has shown the effectiveness of incorporating dual task in semi-supervised NLU. For developing a dialogue system with wide application domains, the domain adaptation and transfer problems of the dual task will be an interesting future research direction.
\end{itemize}

\ifCLASSOPTIONcaptionsoff
  \newpage
\fi



%

\bibliographystyle{IEEEtran}
\bibliography{IEEEabrv,ref}

%

\begin{IEEEbiography}[{\includegraphics[width=1in,height=1.25in,clip,keepaspectratio]{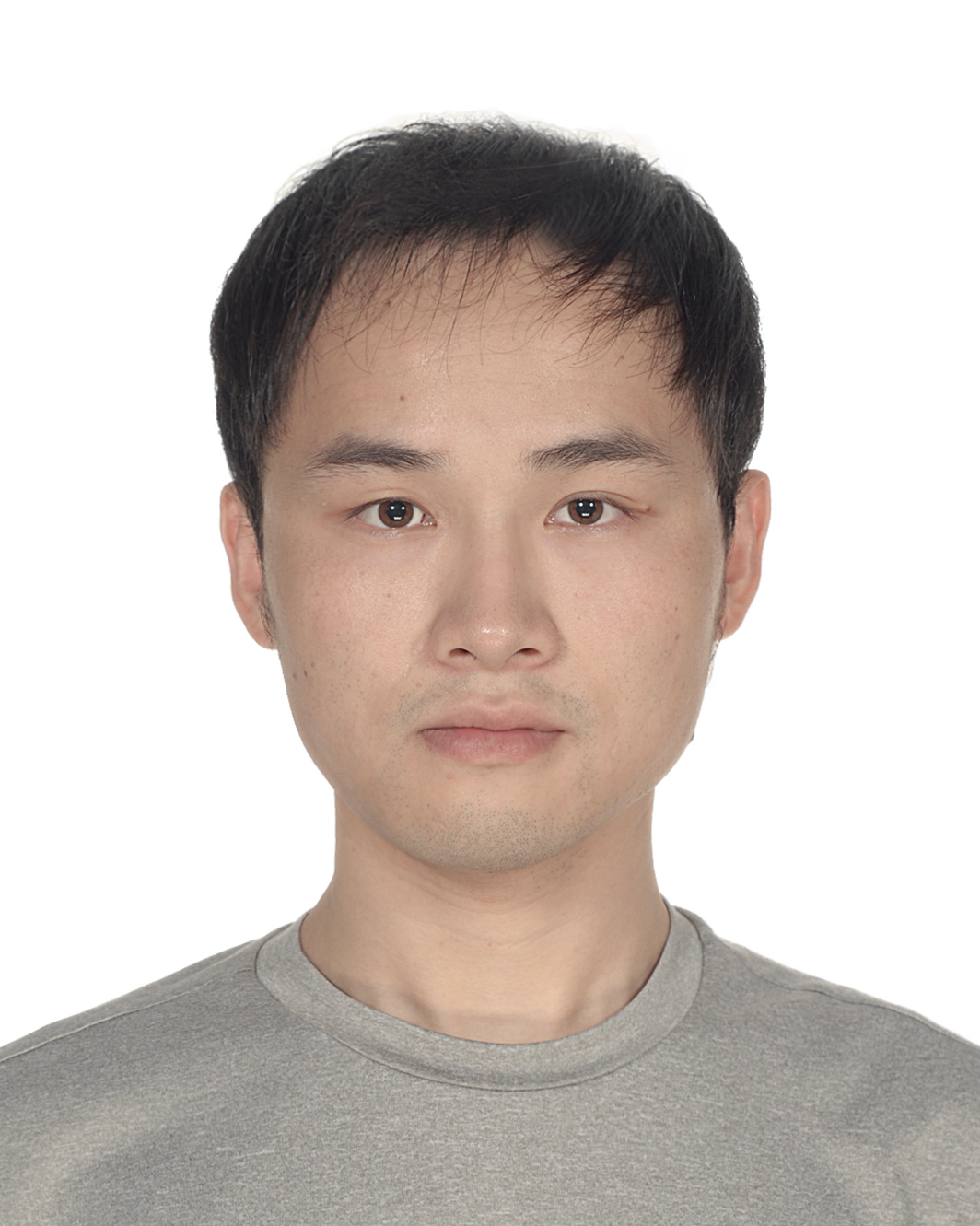}}]{Su Zhu}
received the B.Eng. degree in computer science from Xi’an Jiao Tong University, China in 2013, and the  M.Sc. degree from the Department of Computer Science, Shanghai Jiao Tong University, Shanghai, China, in 2016. He is currently working toward the Ph.D. degree at the SpeechLab, Department of Computer Science and Engineering, Shanghai Jiao Tong University, Shanghai, China. His research interests include spoken/natural language understanding, dialogue systems, and structured deep learning.
\end{IEEEbiography}

\begin{IEEEbiography}[{\includegraphics[width=1in,height=1.25in,clip,keepaspectratio]{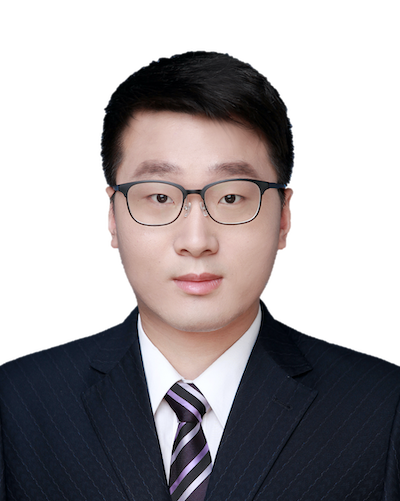}}]{Ruisheng Cao}
received the B.Eng. degree in computer science from Shanghai Jiao Tong University, Shanghai, China, in 2018. He is currently working toward the M.S. degree with the SpeechLab, Department of Computer Science and Engineering, Shanghai Jiao Tong University, Shanghai, China. His research interests include semantic parsing, code generation, spoken language understanding, and machine learning especially structured prediction.
\end{IEEEbiography}


\begin{IEEEbiography}[{\includegraphics[width=1in,height=1.25in,clip,keepaspectratio]{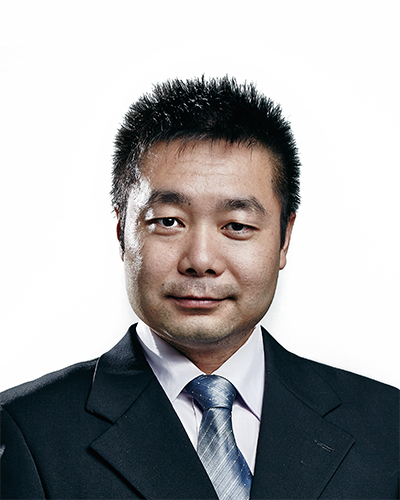}}]{Kai Yu}
is a professor at Computer Science and Engineering Department, Shanghai Jiao Tong University, China. He received his B.Eng. and M.Sc. from Tsinghua University, China in 1999 and 2002, respectively. He then joined the Machine Intelligence Lab at the Engineering Department at Cambridge University, U.K., where he obtained his Ph.D. degree in 2006. His main research interests lie in the area of speech-based human machine interaction including speech recognition, synthesis, language understanding and dialogue management. He is a member of the IEEE Speech and Language Processing Technical Committee.
\end{IEEEbiography}




\end{document}